\documentclass{article}
\PassOptionsToPackage{numbers,compress}{natbib}
\usepackage[preprint]{neurips_2026}
\usepackage[utf8]{inputenc}
\usepackage[T1]{fontenc}
\usepackage{hyperref}
\usepackage{url}
\usepackage{booktabs}
\usepackage{amsfonts}
\usepackage{nicefrac}
\usepackage{microtype}
\usepackage{graphicx}
\usepackage{multirow}
\usepackage{amsmath}
\usepackage{pifont}
\usepackage[table]{xcolor}
\usepackage{xspace}
\newcommand{\cmark}{\textcolor{green!60!black}{\ding{51}}}
\newcommand{\xmark}{\textcolor{red}{\ding{55}}}
\definecolor{prismblue}{RGB}{66, 133, 244}
\definecolor{prismred}{RGB}{219, 68, 55}
\definecolor{prismyellow}{RGB}{244, 180, 0}
\definecolor{prismgreen}{RGB}{15, 157, 88}
\definecolor{prismpurple}{RGB}{155, 39, 176}
\definecolor{colorclosed}{gray}{0.95}
\definecolor{coloropen}{rgb}{0.9, 0.95, 1}
\definecolor{colorlight}{rgb}{1, 0.95, 0.9}
\newcommand{\prism}{%
    \begingroup\textbf{%
      \textcolor{prismblue}{P}%
      \textcolor{prismred}{R}%
      \textcolor{prismyellow}{I}%
      \textcolor{prismgreen}{S}%
      \textcolor{prismpurple}{M}%
    }\endgroup\xspace}
\newcommand{\prismicon}{%
  \raisebox{-0.15\height}{\includegraphics[height=1.6em]{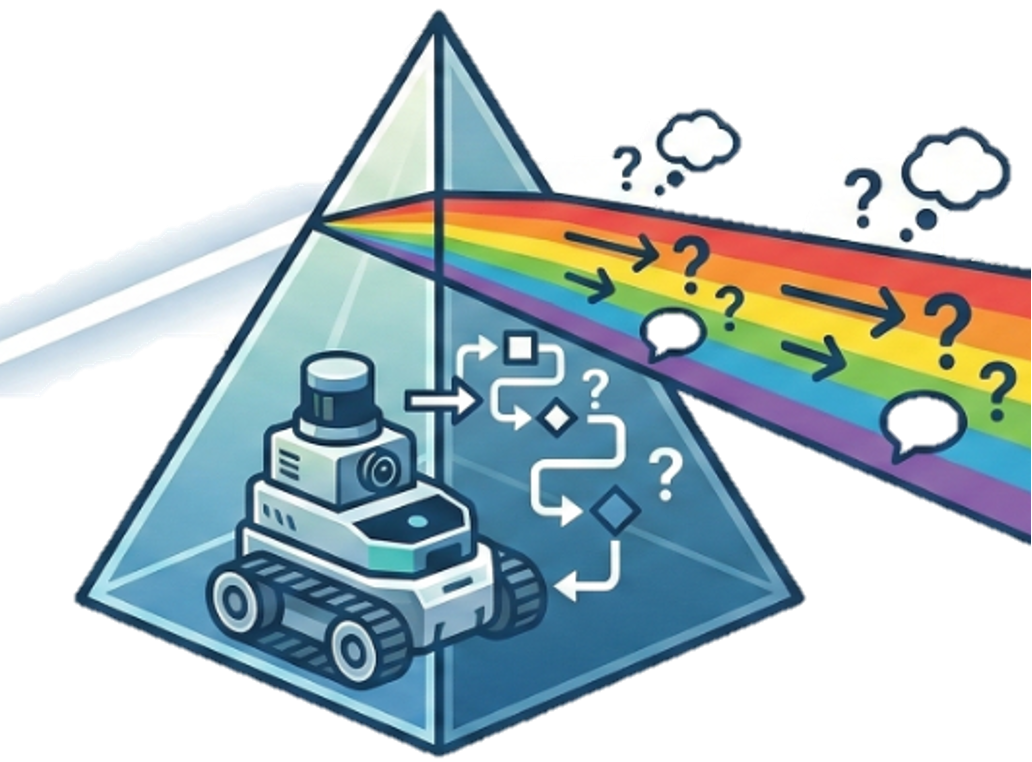}}%
}
\newcommand{\newedit}[1]{#1}

\newcommand{\rev}[1]{\textcolor{red}{#1}}
\title{\protect\prismicon~\prism: Planning and Reasoning with Intent in Simulated Embodied Environments}

\author{
\begin{tabular}{c}
{\bfseries
Yunn Kang Lim$^{1}$,
Pengzhan Sun$^{2}$,
Ziyi Bai$^{3}$,
Xun Xu$^{1}$,} \\
{\bfseries
Angela Yao$^{2}$,
Xulei Yang$^{1}$,
Shijie Li$^{1}$} \\[0.3em]
{\normalfont
$^{1}$A*STAR,
\quad
$^{2}$National University of Singapore,
\quad
$^{3}$BAAI}
\end{tabular}
}

\begin{document}
\maketitle
\begin{abstract}
When an LLM-based embodied agent fails at a household task, the culprit could be misidentified objects, forgotten sub-goals, or poor action sequencing---yet existing benchmarks report only a single success rate, making it impossible to tell which cognitive module is responsible. We present \prism, a diagnostic benchmark that reframes this problem: rather than asking only \textit{did the agent succeed?}, \prism asks \textit{which capability is most likely responsible for failure?} Built on five photorealistic multi-room apartments (4--8 rooms each), \prism structures 300 human-verified tasks into three capability tiers---\textit{Basic Ability}, \textit{Reasoning Ability}, and \textit{Long-horizon Ability}---that isolate perception-to-action grounding, implicit intent resolution, and sustained multi-step coordination respectively. \prism exposes an agent-agnostic executable action API that allows arbitrary agents: LLM agents, VLM agents, symbolic planners, RL policies, and hybrid systems, to be evaluated end-to-end under the same benchmark protocol. To support deeper diagnosis, optional probes for perception, memory, and planning can be adopted, replaced, or bypassed entirely, enabling controlled component-level analysis when desired. Experiments on seven contemporary LLMs establish a clear hierarchy: explicit spatial grounding is not the dominant failure source under oracle perception, implicit intent resolution is a significant bottleneck for all model families, and long-horizon coordination exposes a stark capability cliff---lightweight models collapse to as low as 20.0\% success while simultaneously consuming more tokens than their frontier counterparts, a signature of compensatory over-reasoning rather than genuine planning capability. Project page: \href{https://sj-li.com/PROJ/PRISM}{link}.
\end{abstract}
\section{Introduction}
\label{sec:intro}
A central goal of embodied AI is to build agents that can carry out household tasks from natural-language instructions, such as locating objects, navigating between rooms, and completing multi-step routines. Recent LLMs (Large Language Models) and VLMs (Vision Language Models) have improved the high-level reasoning ability of such agents, but their failures remain difficult to interpret. An unsuccessful episode may be caused by a perceptual mistake, an incorrect use of memory, or a poor sequence of actions. When all of these errors are collapsed into a single task-success score, it becomes unclear which part of the system should be improved.
Most existing embodied-agent benchmarks report performance at the episode level. This is useful for comparing end-to-end systems, but less informative for diagnosing why an agent fails. In particular, perception, memory, and planning errors can lead to the same final outcome, even though they require different remedies. \prism is designed to make these failure sources more separable through a tiered task suite, diagnostic task labels, and optional component-level probes.
\prism is built on five photorealistic Unity~6/HDRP apartment scenes, each containing 4--8 rooms. The benchmark includes 300 human-verified tasks organized into three capability tiers. The \textit{Basic} tier uses explicit instructions to test perception-to-action grounding. The \textit{Reasoning} tier uses vague human-centric utterances, such as \textit{``I'm tired''}, which require the agent to infer an implicit physical goal. The \textit{Long-horizon} tier requires coordinating three or more objects over up to 60 steps, often across multiple rooms. These tiers are not intended only to vary difficulty; they are designed to expose different sources of failure, including grounding errors, intent-resolution errors, and memory-dependent planning errors.
At the benchmark level, \prism only requires agents to interact with the simulator through an executable action API---the same interface for LLM agents, VLM agents, symbolic planners, and RL policies alike. To support deeper diagnosis, we additionally provide an optional reference pipeline with perception, memory, and planning probes that can be adopted, replaced, or bypassed entirely, allowing both LLM-based and non-LLM agents to be evaluated under the same benchmark protocol. In our main experiments, we use oracle perception to isolate reasoning and memory from visual detection errors; additional perception-noise experiments in the appendix show how performance changes under imperfect observations.
Experiments on seven LLMs show a consistent pattern. Under oracle perception, all models achieve more than 66\% success on Basic tasks, {suggesting that explicit spatial grounding is not the dominant failure source in this controlled setting.} Reasoning tasks remain challenging: even the strongest model, GPT-5.2, reaches only 72.2\% success, indicating that implicit intent resolution is still a bottleneck. Long-horizon tasks further separate model families. Lightweight models can drop to 20.0\% success while consuming substantially more tokens than frontier models, suggesting that they often compensate for weak planning or memory use with longer but ineffective reasoning traces.

Our contributions are as follows:
\begin{itemize}
    \item \textbf{A diagnostic benchmark for cross-room long-horizon embodied planning.} \prism's three capability tiers, targeting explicit grounding, implicit intent resolution, and memory-dependent coordination, serve as direct diagnostic signals for per-module failure attribution, enabling controlled diagnosis across the full perception--memory--planning stack.
    \item \textbf{An agent-agnostic evaluation protocol with optional diagnostic probes.} \prism requires only an executable action API, allowing LLM agents, VLM agents, symbolic planners, and RL policies to be evaluated under the same protocol. Optional perception, memory, and planning probes can be independently substituted to isolate specific capability bottlenecks.
    \item \textbf{A scene-graph-driven task generation pipeline} that produces human-verified, apartment-level tasks with explicit cross-room dependencies.
\end{itemize}
\section{Related Work}
\label{sec:related}
\textbf{Embodied Planning Benchmarks.}
AI2-THOR~\cite{kolve2017ai2} and ALFRED~\cite{shridhar2020alfred} established the instruction-following paradigm for household agents, but primarily operate in single-room episodes, limiting evaluation of cross-room spatial memory. VirtualHome~\cite{puig2018virtualhome}, BEHAVIOR-1K~\cite{li2023behavior}, iGibson~2.0~\cite{liigibson}, and Habitat~3.0~\cite{puig2023habitat} introduced multi-room settings and richer physical fidelity. EmbodiedBench~\cite{yang2025embodiedbench} recently scaled evaluation to multimodal LLM agents and introduces six capability subsets; however, it evaluates each capability with separate task sets rather than through a modular pipeline that isolates cognitive components, and its free-form action interface does not structurally prevent action hallucination. ET-Plan-Bench~\cite{zhang2025plan} also evaluates planning ability through structured task categories, but focuses on plan-level evaluation rather than closed-loop apartment-level execution with cross-room dependency labels and optional component-level probes. In contrast, \prism couples apartment-level cross-room episodes with independently substitutable perception, memory, and planning probes, enabling controlled attribution of failures to specific cognitive components.

\newedit{\textbf{Modular Diagnostic Evaluation.}
The Embodied Agent Interface (EAI)~\cite{li2024eai} (NeurIPS 2024 D\&B Oral) decomposes embodied agents into four functional modules and evaluates each with fine-grained metrics on existing single-room environments. \prism is orthogonal: it is a purpose-built \textit{benchmark environment} contributing apartment-level cross-room scenes, a tiered task structure, and an affordance-grounded action space---infrastructure that makes apartment-level diagnostic evaluation possible and that EAI's architectural overlay does not provide.}

\textbf{Hierarchical Memory in Embodied Agents.} 
Long-horizon tasks require retaining persistent spatial knowledge across room transitions while keeping transient execution context up-to-date. Episodic and semantic memory~\cite{park2023generative,zhong2024memorybank}, situation-aware representations~\cite{savva2019habitat,masqa3d}, and architectures that separate persistent from transient context~\cite{zhangbuilding,wangvoyager,packer2023memgpt} all improve long-horizon performance---but evaluate memory solely through end-to-end SR. This conflates two distinct failure modes: an agent that \textit{forgets} sub-goals and one that \textit{remembers} correctly but plans wrongly yield the same SR penalty, obscuring which module needs improvement. KARMA~\cite{wang2025karma} further augments this separation with 3D scene graphs, but still measures memory only through downstream task success. \prism resolves this by instantiating memory as an independently substitutable module, enabling direct attribution of failures to retention versus reasoning at the apartment level.

\textbf{LLM/VLM-Based Embodied Planning.}
LLMs have been integrated as planning backbones via affordance grounding~\cite{brohan2023can}, closed-loop multimodal feedback~\cite{huang2023inner}, and code generation~\cite{liang2023code}; VLMs enable direct visual plan grounding without symbolic state extraction~\cite{driess2023palm,achiam2023gpt}. A persistent confound in evaluating these systems is \textit{action hallucination}: a planner may generate syntactically valid but physically unexecutable actions, and penalizing these silently~\cite{shridhar2020alfred} conflates hallucination failures with genuine reasoning errors. \prism eliminates this confound structurally: the affordance-grounded atomic action space prevents actions outside the predefined executable action space, ensuring that every observed SR gap reflects a real cognitive deficit.

\section{\prism Construction}
\label{sec:benchmark}

\begin{table}[t]
\centering
\caption{Comparison of \prism with representative embodied planning benchmarks. \textit{Implicit Intent}: vague utterances requiring commonsense goal inference. \textit{Memory Probe}: cross-room spatial memory independently evaluable. \textit{Cross-room Diag.}: per-task cross-room dependency labels; $\triangle$~= multi-room scenes supported but dependency not explicitly labeled. \textit{Afford.-Grounded Act.}: action space constrained by affordance ontology. \textit{Optional Probes}: modular components substitutable.}
\label{tab:benchmark_comparison}
\resizebox{\linewidth}{!}{%
\begin{tabular}{lcccccc}
\toprule
\textbf{Benchmark} & \textbf{Planning} & \textbf{Implicit Intent} & \textbf{Memory Probe} & \textbf{Cross-room Diag.} & \textbf{Afford.-Grounded Act.} & \textbf{Optional Probes} \\
\midrule
ALFRED~\cite{shridhar2020alfred}           & \cmark & \xmark & \xmark & \xmark & \xmark & \xmark \\
AI2-THOR~\cite{kolve2017ai2}               & \cmark & \xmark & \xmark & \xmark & \xmark & \xmark \\
VirtualHome~\cite{puig2018virtualhome}     & \cmark & \xmark & \xmark & \rev{$\triangle$} & \xmark & \xmark \\
BEHAVIOR-1K~\cite{li2023behavior}          & \cmark & \xmark & \xmark & \rev{$\triangle$} & \xmark & \xmark \\
EmbodiedBench~\cite{yang2025embodiedbench} & \cmark & \xmark & \xmark & \xmark & \xmark & \xmark \\
\midrule
\textbf{\prism (Ours)} & \cmark & \rev{\cmark} & \rev{\cmark} & \rev{\cmark} & \rev{\cmark} & \rev{\cmark} \\
\bottomrule
\end{tabular}}
\vspace{-8mm}
\end{table}

\textbf{Benchmark infrastructure vs.\ reference agent.} \prism separates two distinct layers. The \textit{benchmark layer}, the simulator, 300 human-verified tasks, affordance-grounded action space, deterministic state evaluator, and per-task diagnostic labels, is architecture-neutral: any agent, whether LLM-based, RL-trained, or rule-based, can be evaluated by issuing the 21 atomic actions through the Python API without adopting any specific internal architecture. The \textit{reference agent layer}, the modular Perception/Memory/Planning pipeline described in Section~\ref{sec:pipeline}, is one concrete instantiation used to establish baselines and validate the diagnostic interfaces via ablation. These two layers are released independently; future work may replace any reference component, or bypass the pipeline entirely and submit actions directly to the environment.

\prism is intentionally designed at a controlled scale to make closed-loop LLM evaluation reproducible and practically affordable. Unlike static QA benchmarks or template-based embodied evaluations, each \prism episode requires repeated simulator interaction and LLM calls to both the Memory and Planning modules. For example, a single GPT-5.2 Long-horizon episode consumes 126.8K tokens on average (Table~\ref{tab:main_results}), and the full benchmark evaluates 300 tasks across 7 models. Increasing the number of apartments would therefore scale evaluation cost nearly linearly and make routine comparison across future methods substantially harder.

Within this cost budget, \prism uses five photorealistic apartments with 4--8 rooms each, yielding approximately 30 room-level spaces embedded in five connected spatial topologies. This design differs from single-room benchmarks such as ALFRED~\cite{shridhar2020alfred} and AI2-THOR~\cite{kolve2017ai2}, where most episodes are localized within a single room and do not test memory across room boundaries as a primary factor. It is also complementary to larger multi-room simulators such as BEHAVIOR-1K~\cite{li2023behavior}: rather than maximizing environment scale, \prism prioritizes controlled diagnostic evaluation, explicitly labeling cross-room dependencies and organizing tasks into capability tiers. The consistent cross-tier trends across all five apartments (Table~\ref{tab:per_apartment}) suggest that the reported bottlenecks are not dominated by any single apartment layout.

\begin{figure}[t]
    \centering
    \includegraphics[width=\linewidth]{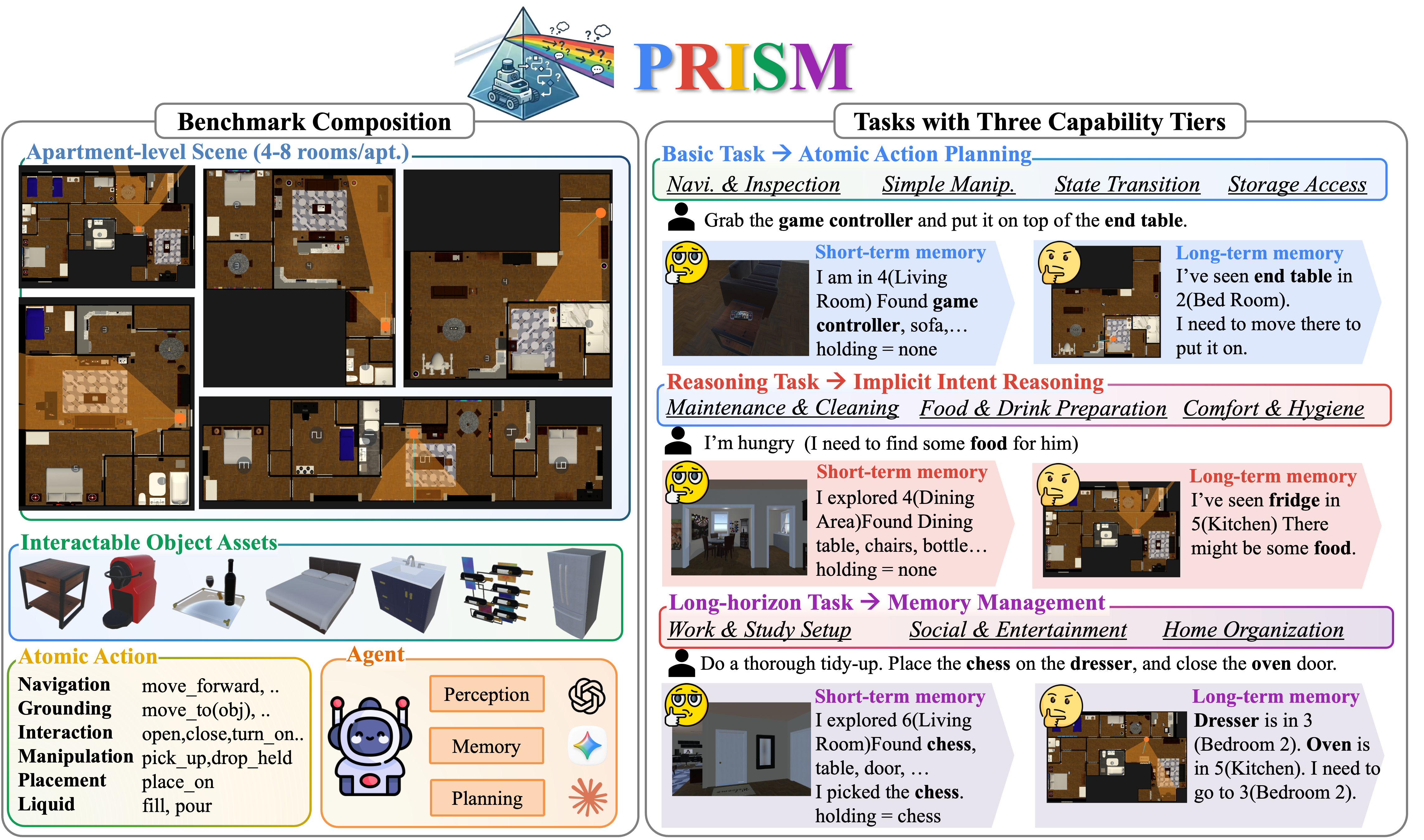}
    \caption{Overview of \prism. The benchmark layer provides apartment-level environments, human-verified tasks, executable actions, deterministic state evaluation, and diagnostic task labels. Optional probes, including oracle perception, memory summaries, target-room prediction, and affordance-grounded action validation, can be enabled for component-level diagnosis but are not required by evaluated agents.}
    \label{fig:teaser}
    \vspace{-4mm}
\end{figure}

\subsection{Preliminaries}
For the optional reference pipeline used in our LLM experiments, we formalize embodied planning as a memory-augmented policy. At each step $t$, the agent receives an egocentric observation $o_t$ and maintains an explicit structured memory state $\mathcal{M}_t = \langle \mathcal{M}_{long}, \mathcal{M}_{short}^{(t)} \rangle$ that separates persistent spatial knowledge from transient execution context:
\begin{equation}
    a_t \sim \pi_{\theta}\!\left(a \mid o_t,\, \mathcal{M}_t,\, \mathcal{I}_{inst}\right), \quad \mathcal{M}_{t+1} = \psi(\mathcal{M}_t, o_{t+1}, a_t).
    \label{eq:policy}
\end{equation}
This two-tier separation is the key design choice: concatenating the raw trajectory history $\{o_0, a_0, \ldots, o_t\}$ instead would cause linear context growth that saturates LLM context windows precisely on the long-horizon tasks where retention matters most. The simulator transition and success evaluator are deterministic and specified by the action preconditions and state deltas in Appendix~\ref{sec:supp_action}; Eq.~\eqref{eq:policy} defines the memory-augmented evaluation interface rather than a learned dynamics model.

\subsection{Benchmark Environment}
\textbf{High-Fidelity Simulator.} 
\prism is implemented in Unity~6/HDRP with photorealistic PBR rendering, physically based lighting, and deterministic state transitions. The simulator supports concurrent evaluation through Unity ML-Agents and exposes a standardized Python API with OpenRouter-compatible endpoints, allowing different LLM backends to be evaluated under the same environment interface. All scene configurations, task definitions, and evaluation scripts will be publicly released; full artifact documentation, license terms, and release contents are described in Appendix~\ref{sec:supp_artifact}.

\textbf{Interactive Scene Construction.}
\prism comprises five apartment scenes, each containing 4--8 rooms with distinct floor-plan topology, room connectivity, object co-occurrence patterns, and cross-room navigation routes. Across these apartments, the benchmark includes 79 semantic object classes and 293 unique 3D assets, covering furniture, appliances, fixtures, and manipulable objects. Objects support binary states such as \texttt{is\_open} and \texttt{is\_on}, as well as relational states such as \texttt{holding}, \texttt{is\_filled}, and \texttt{on\_top\_of}. Each object is assigned an affordance type from a fixed 7-category ontology (Appendix~\ref{sec:supp_affordance}), which defines the legal actions applicable to that object.

\textbf{Executable Affordance-Grounded Action API.}
\prism exposes 21 executable atomic actions grouped into six functional categories: Navigation, Grounding, Interaction, Manipulation, Placement, and Liquid. The purpose of this action API is not to claim a novel action granularity, but to provide a controlled execution interface for diagnostic evaluation. Each action is checked against object state, visibility, room context, and affordance compatibility before execution, so invalid action-format errors are structurally separated from genuine reasoning, memory, or planning failures. This ensures that every observed performance gap across models and tiers reflects a real cognitive difference rather than an interface artifact. Full preconditions and state deltas are provided in Appendix~\ref{sec:supp_action}.

\subsection{Task Design and Capability Tiers}
\label{sec:tasks}
\textbf{Scene-Graph-Driven Task Generation.}
Tasks are generated through a three-stage pipeline: (1) a scene graph is extracted from each apartment, encoding room topology, object placement, object states, and affordance annotations; (2) an LLM generates candidate instructions grounded in that scene graph and constrained to objects and affordances actually present in the apartment; and (3) human annotators execute each candidate task in the simulator to verify completability, instruction clarity, and object reachability. This process yields \textbf{300 human-verified tasks} distributed across five apartments, ten living scenarios, and three capability tiers. Full pipeline details, tier-conditioned generation constraints, and annotator verification criteria are provided in Appendix~\ref{sec:supp_taskgen}.

\textbf{Dataset Control and Verification.}
Beyond scene-graph grounding and human execution checks, we control diagnostic factors by assigning every task to one capability tier and labeling whether it requires cross-room navigation. Reasoning tasks are intentionally single-room, which isolates implicit intent resolution from cross-room memory and navigation, while Basic and Long-horizon tasks include explicit cross-room dependency labels for diagnostic analysis.

\textbf{Capability-Tier Hierarchy.}
\prism organizes tasks hierarchically: each task belongs to one of three top-level \textit{capability tiers}, each tier contains several living scenarios, and each scenario contains concrete executable tasks. The three tiers are designed to emphasize different sources of failure rather than merely increasing task difficulty: \textit{Basic Ability} targets explicit grounding, \textit{Reasoning Ability} targets implicit intent resolution, and \textit{Long-horizon Ability} targets memory-dependent multi-step coordination. The detailed capability definitions are shown in Appendix~\ref{app:capability}.

\textbf{Cross-Room Labels as an Orthogonal Diagnostic Attribute.}
In addition to its capability tier, each task is labeled according to whether successful completion requires crossing room boundaries. The verified cross-room proportions are: Basic 38.3\% (46/120), Reasoning 0\% (all single-room by design), and Long-horizon 45.6\% (41/90), with the full distribution reported in Appendix~\ref{sec:supp_dataset}. Unlike prior multi-room simulators where cross-room dependence is not isolated as a diagnostic factor, \prism explicitly exposes this property for every task. Cross-room tasks are measurably harder: single-room vs.\ cross-room SR differs by 18.9 pp on Basic tasks (86.1\% vs.\ 67.2\%) and 9.9 pp on Long-horizon tasks (68.1\% vs.\ 58.2\%), confirming that room-boundary transitions add genuine cognitive pressure.

\section{Agent-Agnostic Evaluation Protocol and Diagnostic Probes}
\label{sec:pipeline}

At the benchmark level, \prism only requires an agent to receive egocentric observations and issue executable actions through the environment API. This protocol is architecture-neutral: LLM agents, VLM agents, symbolic planners, RL policies, and hybrid systems can all be evaluated without any particular internal architecture. Success is determined solely by whether the agent reaches the goal state within the step budget, measured by the deterministic state evaluator.

For component-level diagnosis, \prism additionally provides an \textbf{optional reference pipeline} that decomposes agent behavior into three independently substitutable probes---Perception, Memory, and Planning---connected through standardized interfaces. Using these probes is not required to run the benchmark; they exist to help researchers isolate \textit{which} cognitive component is responsible for observed failures. Memory and planning probe substitution is validated via controlled ablations (Tables~\ref{tab:ablation_memory}--\ref{tab:ablation_planning}); the perception probe is characterized via noise experiments (Appendix~\ref{sec:supp_ablation}). The oracle Perception probe removes visual noise to isolate reasoning; the affordance-grounded action interface eliminates hallucination artifacts; and the hierarchical Memory probe (Historical Summary + Target Room Prediction) provides a minimal swappable baseline against which any memory design can be benchmarked.

\begin{figure}[t]
    \centering
    \includegraphics[width=\linewidth]{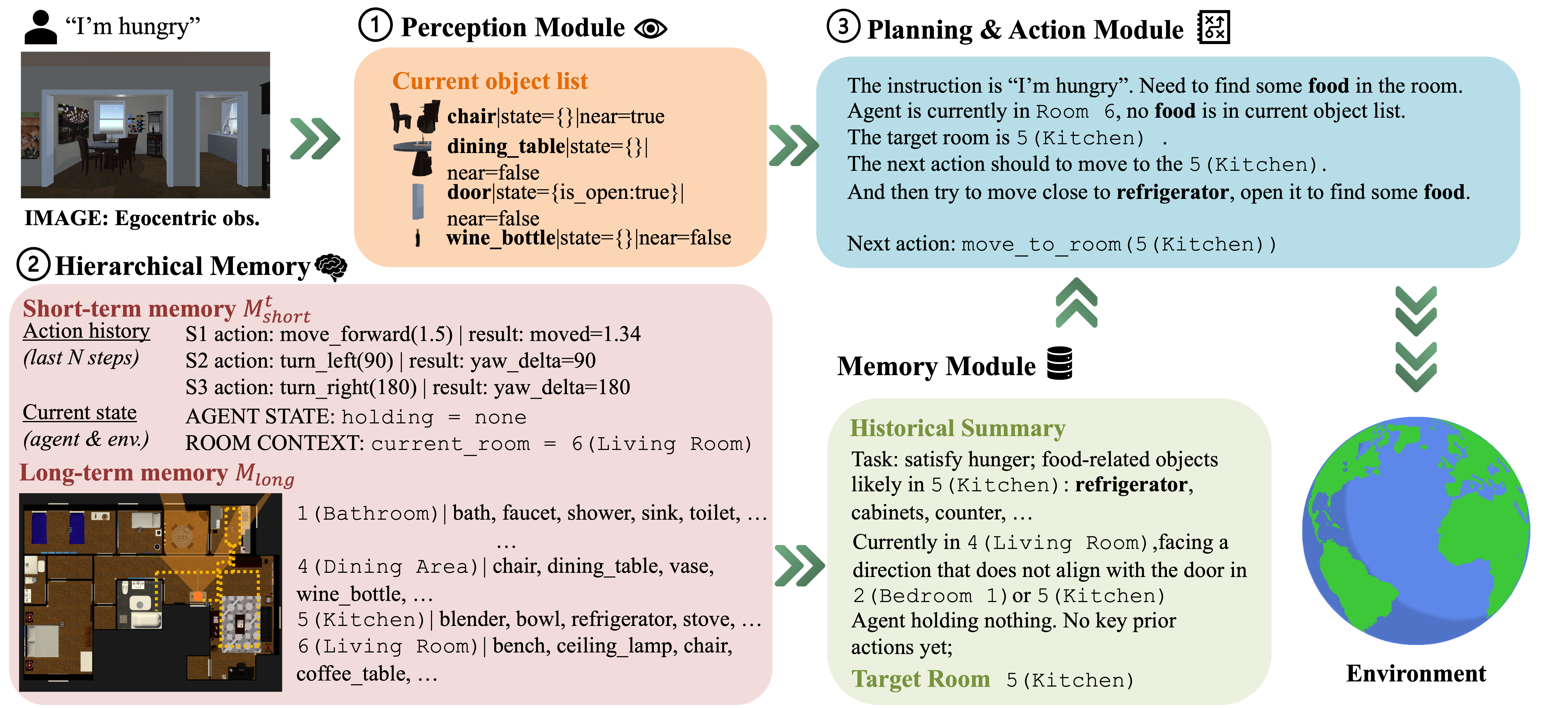}
    \caption{The \prism reference diagnostic pipeline (optional). At each step, the Perception probe produces a structured object list from the egocentric observation; the Memory probe synthesizes it with persistent spatial knowledge and execution history into a Historical Summary and Target Room Prediction; the Planning probe selects a single affordance-constrained atomic action. Each probe can be independently replaced to measure its isolated contribution. State deltas merge back into memory.}
    \label{fig:planning}
    \vspace{-4mm}
\end{figure}

The modules described below constitute the \textit{reference agent}---one specific instantiation used to establish baselines and validate the diagnostic interfaces, not a mandatory architecture for evaluated agents. The benchmark infrastructure (simulator, tasks, action space, evaluator) is fully decoupled from this reference implementation. An RL-trained policy, a classical planner, or a non-LLM agent can be evaluated by issuing actions through the same API without implementing any of the modules below. Researchers who \textit{do} adopt the modular interfaces gain the ability to swap individual components and measure their isolated contribution against the common baseline reported here---but this is an option, not a requirement.

\subsection{Perception Module}
The Perception module maps the current egocentric observation to a structured list of visible object types $\mathcal{P}_t$, providing the standardized interface between raw visual input and the downstream memory and planning modules. Any perception model---from a lightweight object detector to a frontier VLM---can be substituted at this interface without modifying downstream modules, enabling controlled evaluation of how perception quality propagates into planning performance.

In our experiments, we adopt \textbf{oracle} configuration, which provides ground-truth visible object lists directly from  simulator. This is a deliberate experimental choice that isolates reasoning and memory capability from perceptual noise, establishing a clean upper-bound reference for each model's cognitive performance. It is not an architectural limitation of  benchmark, the interface is explicitly designed for substitution. On  45-task ablation subset with GPT-5.2, controlled perception noise yields a gradual decrease from 74.0\% SR under oracle perception to 72.0\% with 20\% object drop and 64.0\% with 40\% object drop (full table in Appendix~\ref{sec:supp_ablation}), with the largest degradation on Long-horizon tasks where missed objects compound across sub-goals. We additionally evaluate a VLM-based perception condition in which a frontier VLM identifies visible object types from  egocentric image, mapped to oracle object IDs; this yields a larger SR drop to 62.0\%, worse than  40\% random drop condition, driven by systematic rather than stochastic detection errors (full analysis in Appendix~\ref{sec:supp_ablation}).

\subsection{Reference Memory Probe}
\label{subsec:memory}
The reference diagnostic pipeline includes an optional memory probe that separates persistent spatial knowledge from transient execution context. This probe is not required by the benchmark; it is used in our experiments to study whether failures arise from missing spatial knowledge (long-term), poor history compression (short-term), or incorrect room-level navigation decisions. Any alternative memory design, including a 3D scene graph~\newedit{\cite{wang2025karma}}, a learned consolidation model, or a graph-based spatial reasoner, can replace this probe while keeping the rest of the pipeline fixed, yielding a directly comparable measurement. The detailed memory definitions are shown in Appendix~\ref{app:memory}.

Rather than forwarding logs to planner, Memory probe synthesizes these two streams into a \textbf{Historical Summary}, a free-form consolidation of completed sub-goals and pending actions that prevents redundant re-exploration, and a \textbf{Target Room Prediction} that infers next navigation destination from task progress and long-term spatial knowledge, reducing aimless room-by-room search.

\subsection{Affordance-Grounded Action Selection}
\label{subsec:planning}
The Planning module synthesizes the task instruction $\mathcal{I}_{inst}$ with the Memory module's outputs to select exactly one executable action per step. To structurally eliminate action hallucination, target selection is restricted to objects with three jointly verified properties: (1) a \textit{room context} match (the object is in a room accessible from the current position); (2) a \textit{perception match} (the object appears in the current visible object list with a unique ID); and (3) an \textit{affordance match} (the object's affordance type is compatible with the selected action, e.g., \texttt{open} can only target objects with \texttt{door} or \texttt{container} affordance). This three-way grounding ensures that every observed performance difference across models and tiers reflects a genuine reasoning capability gap rather than an interface-level execution artifact---a prerequisite for the diagnostic conclusions in Section~\ref{sec:experiments}. Full prompt templates for both the Memory and Planning modules are provided in Appendix~\ref{sec:supp_prompt}.

\subsection{Evaluation Metrics}
\label{subsec:metrics}
\prism reports three complementary metrics per tier: \textbf{Success Rate (SR)}, the fraction of tasks in which the agent reaches the goal state within the 60-step budget; \textbf{Steps}, the average number of actions taken to complete successful tasks; and \textbf{Token Count (TC)}, the total LLM tokens consumed across all steps. Reporting SR, Steps, and TC simultaneously is important: as we show in Section~\ref{sec:experiments}, smaller models sometimes \textit{increase} token consumption while decreasing SR, a pattern that is invisible in SR-only evaluations and reveals a qualitatively distinct failure mode.

\section{Experiments}
\label{sec:experiments}
We evaluate seven models spanning three categories. On the \textit{proprietary frontier} side: \textbf{GPT-5.2}~\cite{singh2025openai}, \textbf{Claude~Sonnet~4.6}~\cite{anthropic2025claude46}, and \textbf{Gemini~3~Flash}~\cite{google2025gemini3}. On the \textit{open-source frontier}: \textbf{GLM-4.6V}~\cite{glm2024chatglm}. On the \textit{lightweight} side: \textbf{GPT-5~mini}, \textbf{Claude~Haiku~4.5}, and \textbf{Gemini~2.5~Flash~Lite}---covering the cost-performance frontier of each proprietary family. All experiments use oracle perception and identical prompts across models. Full hyperparameters (memory window size, interaction range, temperature, parallel instance count) are documented in Appendix~\ref{sec:supp_impl}.

\subsection{Main Results}
Table~\ref{tab:main_results} reports the full results. We discuss each tier in turn, then characterize the cross-tier pattern.

\begin{table}[t]
\centering
\caption{Performance of frontier LLM agents on \prism across three capability tiers. \colorbox{colorclosed}{Proprietary}, \colorbox{coloropen}{Open-source}, \colorbox{colorlight}{Lightweight}. SR (\%,$\uparrow$), Steps ($\downarrow$), TC ($\times10^3$, $\downarrow$). Best value per column in \textbf{bold}.}
\label{tab:main_results}
\resizebox{\linewidth}{!}{%
\begin{tabular}{l ccc ccc ccc c}
\toprule
& \multicolumn{3}{c}{\textbf{Basic} (120 tasks)}
& \multicolumn{3}{c}{\textbf{Reasoning} (90 tasks)}
& \multicolumn{3}{c}{\textbf{Long-horizon} (90 tasks)}
& \textbf{Overall} \\
\cmidrule(lr){2-4}\cmidrule(lr){5-7}\cmidrule(lr){8-10}\cmidrule(lr){11-11}
\textbf{Model} & SR & Steps & TC & SR & Steps & TC & SR & Steps & TC & SR \\
\midrule
\rowcolor{colorclosed} GPT-5.2
    & \textbf{91.7} & \textbf{14.4} & \textbf{58.4} & \textbf{72.2} & 23.7 & \textbf{100.6} & 80.0 & 30.0 & 126.8 & \textbf{82.3} \\
\rowcolor{colorclosed} Claude Sonnet~4.6
    & 85.0 & 15.2 & 67.5 & 71.1 & \textbf{22.3} & 102.3 & \textbf{82.2} & \textbf{25.9} & 117.2 & 80.0 \\
\rowcolor{colorclosed} Gemini~3 Flash
    & 75.8 & 21.2 & 80.4 & 65.6 & 26.7 & 102.5 & 74.4 & 29.5 & \textbf{113.1} & 72.3 \\
\hline
\rowcolor{colorlight} GPT-5~mini
    & 77.5 & 21.1 & 108.2 & 68.9 & 26.3 & 136.5 & 65.6 & 35.7 & 190.3 & 71.3 \\
\rowcolor{colorlight} Claude Haiku~4.5
    & 81.7 & 20.1 & 84.3 & 68.9 & 26.3 & 112.8 & 64.4 & 37.0 & 159.1 & 72.7 \\
\rowcolor{colorlight} Gemini~2.5 Flash Lite
    & 66.7 & 26.0 & 118.1 & 70.0 & 25.0 & 111.7 & 20.0 & 51.6 & 245.2 & 53.7 \\
\hline
\rowcolor{coloropen} GLM-4.6V
    & 79.2 & 19.6 & 95.6 & 68.9 & 27.8 & 133.5 & 57.8 & 38.5 & 194.3 & 69.7 \\
\bottomrule
\end{tabular}}
\end{table}

\noindent\textbf{Basic Ability (RQ1).}
All seven models exceed 66\% SR, {suggesting that explicit spatial grounding is not the dominant failure source under oracle perception in \prism's Basic tier.} GPT-5.2 leads at 91.7\% with only 14.4 average steps, reflecting efficient single-room planning. Notably, Claude Haiku~4.5 (81.7\%) outperforms Gemini~3~Flash (75.8\%) despite being a lightweight model, suggesting that instruction-following fluency rather than raw reasoning capacity determines Basic-tier outcomes.

\noindent\textbf{Reasoning Ability (RQ2).}
Replacing explicit commands with vague human-centric utterances induces a consistent SR drop across all models, but the magnitude and mechanism differ. GPT-5.2 leads at 72.2\%, while the remaining six models cluster between 65.6\% and 71.1\%. Two failure modes dominate in our qualitative analysis: \textit{semantic hallucination}, where the model navigates to a plausible but contextually wrong destination (e.g., a sofa rather than a bed in response to ``I'm tired''), and \textit{exploratory deadlock}, where the agent enters a repetitive navigation loop exhausting the step budget (see Appendix~\ref{sec:supp_failures} for full taxonomy). Lightweight models cluster near 68--70\% but at substantially higher TC than frontier counterparts (up to 136.5K vs.\ 100.6K for GPT-5.2), a pattern indicating that weaker commonsense grounding is being compensated by more verbose, exploratory reasoning traces rather than more accurate intent resolution.

\noindent\textbf{Long-horizon Ability (RQ3).}
The Long-horizon tier produces the benchmark's most differentiated results and its most informative failure signatures. Claude Sonnet~4.6 leads at 82.2\% with 25.9 average steps---\textit{fewer} steps than GPT-5.2's 80.0\% at 30.0 steps---suggesting that Claude's advantage on this tier is not merely SR but more efficient sub-goal sequencing under sustained planning pressure. The sharpest result is Gemini~2.5-Flash-Lite's collapse to 20.0\% SR, accompanied by a TC of 245.2K against Gemini~3~Flash's 113.1K at 74.4\% SR. This 2.2$\times$ token differential on a 54.4 pp SR gap is the clearest expression of the \textit{context saturation} failure mode: the model completes early sub-goals correctly but as the context window fills, later sub-goal instructions are displaced, causing coherent progress to degrade into repetitive interaction attempts at already-completed states.

\noindent\textbf{Cross-tier Pattern.}
The most striking cross-tier observation is the \textit{inverse SR--TC relationship} in lightweight models: as SR falls from Basic to Long-horizon, TC rises rather than falls. Gemini~2.5-Flash-Lite's TC increases from 118.1K (Basic) to 245.2K (Long-horizon) while its SR falls from 66.7\% to 20.0\%. This is not simply a consequence of longer episodes---successful frontier models consume far fewer tokens on comparable tasks. The pattern instead reflects compensatory over-reasoning: lightweight models generate longer exploratory reasoning traces in an attempt to recover from weaker commonsense grounding and memory retention, ultimately exhausting both the step and token budgets without recovering task success. This SR--TC inversion is only visible because \prism reports both metrics jointly; it would be invisible under SR-only evaluation.

\subsection{Per-Apartment Generalization}
A natural concern with a five-scene benchmark is whether findings are robust to scene-specific layout choices. Table~\ref{tab:per_apartment} addresses this directly. GPT-5.2's cross-tier ordering (Basic $>$ Long-horizon $\geq$ Reasoning) is consistent across all five apartments, and the per-apartment overall SR standard deviation is only 2.5\%. The consistent pattern---not the absolute values---is the key evidence: the cognitive bottlenecks identified by \prism are a property of the task tiers, not artifacts of specific room configurations.

\begin{table}[t]
\centering
\caption{GPT-5.2 SR (\%) disaggregated by apartment and capability tier. The cross-tier ordering is consistent across all five distinct floor-plan topologies (overall SR std = 2.5\%), indicating that benchmark conclusions reflect cognitive capability differences rather than scene-specific layout.}
\label{tab:per_apartment}
\resizebox{0.8\linewidth}{!}{%
\begin{tabular}{lcccc}
\toprule
\textbf{Apartment} & \textbf{Basic SR} & \textbf{Reasoning SR} & \textbf{Long-horizon SR} & \textbf{Overall SR} \\
\midrule
Apartment 1 & 91.7 & 66.7 & 83.3 & 80.6 \\
Apartment 2 & 100.0 & 72.2 & 83.3 & 85.2 \\
Apartment 3 & 87.5 & 83.3 & 75.0 & 82.1 \\
Apartment 4 & 95.8 & 66.7 & 72.2 & 78.2 \\
Apartment 5 & 83.3 & 72.2 & 83.3 & 79.6 \\
\midrule
Mean $\pm$ Std & 91.7$\pm$6.0 & 72.2$\pm$6.7 & 79.4$\pm$5.0 & 81.1$\pm$2.5 \\
\bottomrule
\end{tabular}}
\vspace{-4mm}
\end{table}

\subsection{Ablation Studies}
\label{sec:ablation}
We isolate the contribution of each Memory and Planning module component using GPT-5.2 on a 45-task ablation subset (15 tasks per tier), holding all other variables fixed. For reference, a human baseline on the same subset reaches 92.0\% SR with 15.3 average steps (Appendix~\ref{sec:supp_ablation}), leaving an 18-point gap to the GPT-5.2 reference agent.

\noindent\textbf{Memory Module} (Table~\ref{tab:ablation_memory}).
Removing textual long-term memory (M2) produces the largest SR drop at $-$14\%, concentrated disproportionately on Reasoning tasks where object-level room semantics are essential for implicit intent resolution: without knowing which room contains a \texttt{dining\_table}, the Memory module cannot generate a meaningful Target Room Prediction for \textit{``I'm hungry''}, leading to topologically plausible but semantically mismatched navigation. More surprisingly, \textit{adding} a top-down visual map (M3) \textit{decreases} SR by 2\% while adding 41.2K tokens---a counterintuitive result explained by the map introducing geometric proximity signals that occasionally override the more semantically reliable textual room inventory, directing the agent toward the geometrically nearest room rather than the semantically appropriate one.

\begin{table}[t]
\centering
\caption{Ablation of Memory module inputs. LTM: Textual Long-term Memory. Visual Map: top-down floor plan image.}
\label{tab:ablation_memory}
\resizebox{0.78\linewidth}{!}{%
\begin{tabular}{cl cc ccc}
\toprule
\textbf{Exp} & \textbf{Configuration} & \textbf{LTM} & \textbf{Visual Map} & \textbf{SR\,$\uparrow$} & \textbf{Steps\,$\downarrow$} & \textbf{TC ($\times10^3$)\,$\downarrow$} \\
\midrule
M1 & \textbf{Full (Ours)}   & \cmark & \xmark & \textbf{74.0} & \textbf{26.6} & \textbf{113.1} \\
M2 & w/o LTM               & \xmark & \xmark & 60.0 & 31.5 & 119.6 \\
M3 & with Visual Map        & \cmark & \cmark & 72.0 & 26.5 & 154.3 \\
\bottomrule
\end{tabular}}
\vspace{-4mm}
\end{table}

\noindent\textbf{Planning Module} (Table~\ref{tab:ablation_planning}).
Historical Summary is  dominant Planning component: removing it (P2) drops SR by 36\% and inflates Steps to 39.7, as  planner, deprived of a consolidated progress record, repeatedly re-navigates to already-completed states. Replacing the summary with raw action-history concatenation (P3) partially recovers to 62.0\% but degrades steeply beyond ${\sim}25$ steps as context saturation sets in, while adding 17.9K tokens per episode, confirming that the value of the Historical Summary lies in \textit{compression} rather than \textit{retention}. Removing Target Room Prediction (P4) produces a modest $-$8\% SR drop, but failures concentrate on Long-horizon tasks where  planner, lacking an explicit destination, reverts to reactive room-by-room exploration and accumulates navigation overhead that compounds across multi-object chains. Removing Room Options (P5) causes only $-$2\% SR drop but reduces TC by 6.8K, as the planner occasionally generates non-existent room names in later sub-goals without grounding constraint, a subtle but consistent source of silent navigation failures in Long-horizon tasks where the step budget is too depleted to recover.

\begin{table}[t]
\centering
\caption{Ablation of Planning module inputs. HS: Historical Summary. TR: Target Room Prediction. RO: Room Options.}
\label{tab:ablation_planning}
\resizebox{0.9\linewidth}{!}{%
\begin{tabular}{cl ccc ccc}
\toprule
\textbf{Exp} & \textbf{Configuration} & \textbf{HS} & \textbf{TR} & \textbf{RO} & \textbf{SR\,$\uparrow$} & \textbf{Steps\,$\downarrow$} & \textbf{TC ($\times10^3$)\,$\downarrow$} \\
\midrule
P1 & \textbf{Full (Ours)}          & \cmark     & \cmark & \cmark & \textbf{74.0} & \textbf{26.6} & \textbf{113.1} \\
P2 & w/o Historical Summary        & \xmark     & \cmark & \cmark & 38.0 & 39.7 & 154.2 \\
P3 & Raw history (replace Summary) & (raw seq.) & \cmark & \cmark & 62.0 & 29.7 & 131.1 \\
P4 & w/o Target Room               & \cmark     & \xmark & \cmark & 66.0 & 28.5 & 115.8 \\
P5 & w/o Room Options              & \cmark     & \cmark & \xmark & 72.0 & 26.5 & 106.3 \\
\bottomrule
\end{tabular}}
\vspace{-4mm}
\end{table}

\section{Conclusion}
\label{sec:conclusion}
We presented \prism, a diagnostic benchmark that helps separate embodied planning failures across perception, memory, and planning components via an agent-agnostic executable action API, capability-tiered diagnostic task labels, and optional perception, memory, and planning probes. Experiments on seven contemporary LLMs show that {explicit spatial grounding is not the dominant failure source under oracle perception}, implicit intent resolution is a structural capability gap (no model exceeds 73\% on Reasoning), and lightweight models exhibit a characteristic SR--TC inversion on Long-horizon tasks that reveals compensatory over-reasoning. Ablations identify the Historical Summary and object-level long-term memory as the most critical components, and confirm that information compression quality matters more than quantity. The modular interfaces support direct substitution of any component---oracle perception, the textual room inventory~\newedit{\cite{wang2025karma}}, or the minimal planner---with each substitution yielding a directly interpretable measurement against the common baseline reported here. \textbf{Limitations:} current experiments use oracle perception (noise experiments in Appendix~\ref{sec:supp_ablation}); scenes cover household environments only; LLM-generated tasks may carry generator phrasing biases; 3D assets are licensed for academic use only. Extended discussion of limitations and responsible use is provided in Appendix~\ref{sec:supp_limitations}.

\bibliographystyle{abbrvnat}
\bibliography{main}

\appendix

\begin{figure}[t]
    \centering
    \includegraphics[width=\linewidth]{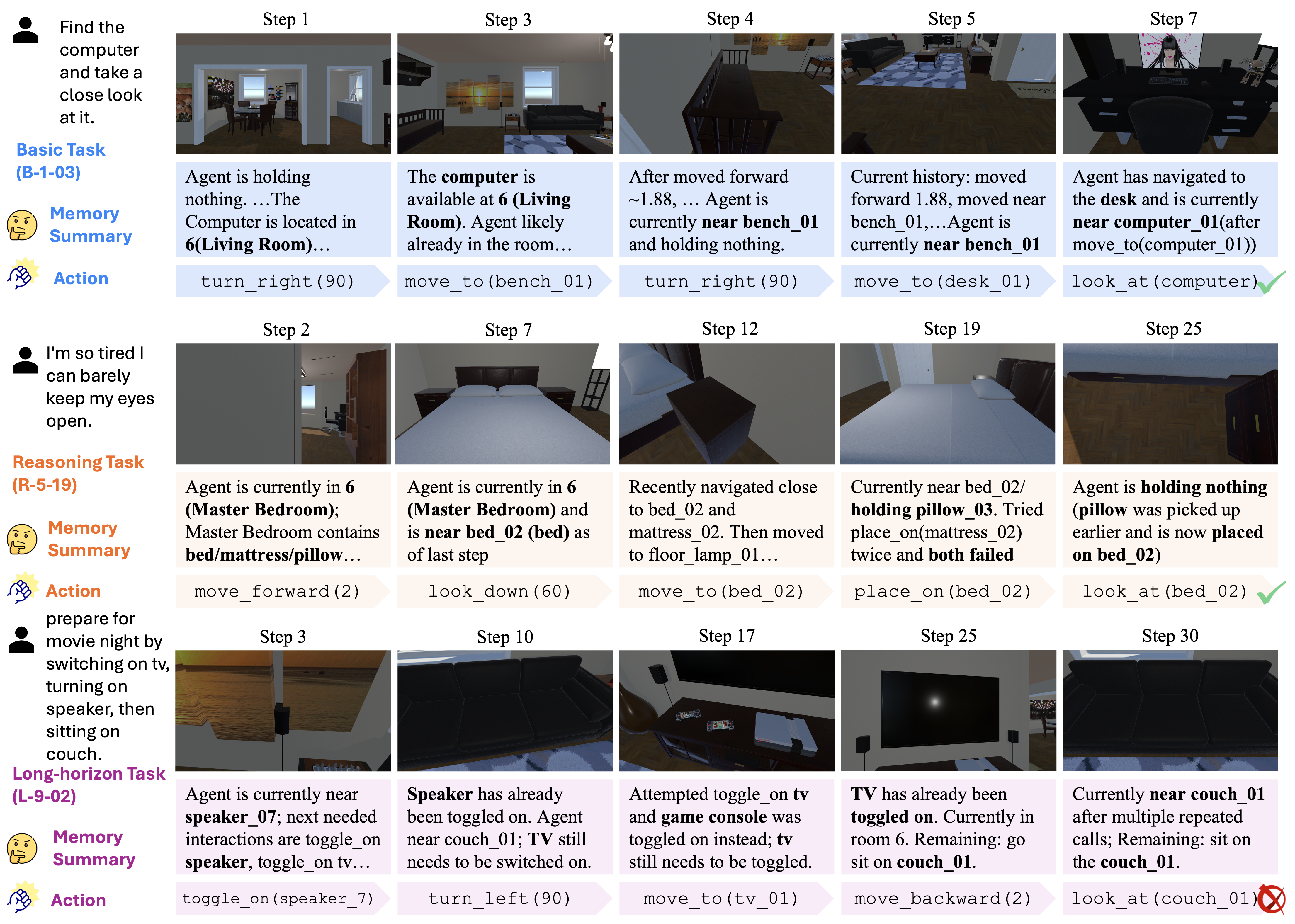}
    \caption{Step-by-step execution traces for GPT-5.2 on representative tasks from each tier, showing egocentric observations, Historical Summary snapshots, and selected actions. The Basic and Reasoning tasks are completed within budget; the Long-horizon task illustrates context saturation collapse---early sub-goals succeed but coherent progress degrades as trajectory context fills the context window.}
    \label{fig:exp}
    \vspace{-5mm}
\end{figure}

\section{Dataset Diversity and Scalability}
\label{sec:supp_dataset}

\subsection{Apartment-Level Scene Design}

\prism comprises five distinct apartment scenes, each featuring an independent floor-plan topology ranging from 4 to 8 rooms. This creates a controlled gradient of spatial complexity: from compact two-bedroom layouts to larger apartments with intricate multi-zone connectivity. Core functional areas include bedrooms, bathrooms, kitchens, and living rooms; auxiliary spaces (studies, dining areas, utility rooms) vary by apartment, ensuring no two apartments share identical room sequences.

This apartment-scale design generates environmental diversity along two axes: (1)~\textit{spatial reasoning diversity}, where differing room layouts require qualitatively different navigation routes for structurally similar task instructions; and (2)~\textit{contextual reasoning diversity}, where identical object categories appear under different surrounding contexts, preventing agents from exploiting layout-specific shortcuts.

\subsection{Object Coverage and State Representation}

Across the five apartments, \prism contains \textbf{79 semantic object classes} and \textbf{293 distinct 3D assets}, spanning furniture (beds, tables, dressers, chairs, sofas), appliances (refrigerators, ovens, stoves, computers, televisions, speakers), fixtures (faucets, showers, toilets, lamps), and manipulable items (bottles, bowls, pillows, chess pieces, keyboards, mice, game controllers, etc.).

Each object exposes a typed physical state: \textit{binary toggles} (\texttt{is\_open}, \texttt{is\_on}) and \textit{relational states} (\texttt{holding}, \texttt{on\_top\_of}, \texttt{is\_filled}). These state transitions serve as ground truth for task success evaluation, ensuring the benchmark measures causal manipulation over time rather than navigation proximity alone.

\subsection{Task Scale and Cross-Room Distribution}

The 300 verified tasks are uniformly distributed across ten living scenarios (30 tasks per scenario), five apartments, three capability tiers, and 67 interacted object categories. Table~\ref{tab:crossroom_dist} summarizes the cross-room distribution per tier.

\begin{table}[h]
\centering
\caption{Cross-room task distribution across capability tiers (300 evaluated tasks).}
\label{tab:crossroom_dist}
\resizebox{0.65\linewidth}{!}{%
\begin{tabular}{lccc}
\toprule
\textbf{Tier} & \textbf{Total} & \textbf{Cross-room} & \textbf{Cross-room \%} \\
\midrule
Basic Ability         & 120 & 46 & 38.3\% \\
Reasoning Ability     & 90  & 0  & 0.0\%  \\
Long-Horizon Ability  & 90  & 41 & 45.6\% \\
\midrule
\textbf{All}          & \textbf{300} & \textbf{87} & \textbf{29.0\%} \\
\bottomrule
\end{tabular}}
\end{table}

Notably, Long-horizon tasks have \textbf{45.6\% cross-room} tasks, which is higher than the 20--30\% design target used in earlier drafts and reflects the verified dataset distribution. Cross-room tasks are more challenging: on Basic tasks, agents achieve 86.1\% SR on single-room vs.\ 67.2\% on cross-room tasks, and on Long-horizon tasks, 68.1\% vs.\ 58.2\%, confirming that cross-room navigation imposes genuine additional cognitive demand.

\section{Task Generation Pipeline and Quality Control}
\label{sec:supp_taskgen}

\subsection{Pipeline Overview}

\prism adopts a three-stage pipeline followed by a human quality-control gate:

\begin{center}
\small
\textbf{Stage 1: Scene Graph Extraction} $\rightarrow$ \textbf{Stage 2: LLM-Based Task Generation} $\rightarrow$ \textbf{Stage 3: Human Feasibility Verification}
\end{center}

\subsection{Stage 1: Scene Graph Extraction}

For each apartment, a \textbf{scene graph} is automatically extracted from the simulator state, encoding: room nodes with spatial adjacency (shared doorways); object nodes with semantic class, affordance type, and physical state; placement edges (object $\to$ room); and connectivity edges (doorway adjacency). This structurally prevents generating instructions that are semantically plausible but physically infeasible.

\subsection{Stage 2: LLM-Based Task Generation}

A frontier LLM is prompted with the scene graph and \textbf{tier-conditioned constraints} across ten predefined living scenarios (Table~\ref{tab:scenario_tiers}):

\begin{table}[h]
\centering
\caption{Living scenarios and their associated capability tiers.}
\label{tab:scenario_tiers}
\resizebox{0.5\linewidth}{!}{%
\begin{tabular}{clc}
\toprule
\textbf{ID} & \textbf{Scenario} & \textbf{Tier} \\
\midrule
1  & Navigation \& Inspection  & Basic \\
2  & Simple Manipulation       & Basic \\
3  & Appliance Control         & Basic \\
4  & Storage Access            & Basic \\
5  & Hygiene \& Comfort        & Reasoning \\
6  & Food \& Drink             & Reasoning \\
7  & Environmental Regulation  & Reasoning \\
8  & Study \& Work Setup       & Long-horizon \\
9  & Entertainment Setup       & Long-horizon \\
10 & Home Organization         & Long-horizon \\
\bottomrule
\end{tabular}}
\end{table}

Tier-specific constraints: \textit{Basic}: fully explicit instructions, named objects, 1--2 objects. \textit{Reasoning}: human-centric utterances with no direct object mention; inferable via commonsense. \textit{Long-horizon}: 3+ distinct objects (mean 3.72/task, range 3--5), $\geq$2 state-changing interactions.

\subsection{Stage 3: Human Feasibility Verification}

All generated tasks undergo \textbf{mandatory human feasibility verification}: annotators execute each task directly in the simulator and assess (1) Executability within 60 steps, (2) Instruction Clarity at the intended tier, and (3) Object Reachability. Tasks failing any criterion are revised or discarded. Cross-room tasks are additionally labeled with required room boundary transitions.

\section{Three Tiers Capability}
\label{app:capability}
\begin{itemize}
    \item \textbf{Tier 1: Basic Ability (120 tasks).}
The Basic tier contains four scenarios: \textit{Navigation \& Inspection}, \textit{Simple Manipulation}, \textit{Appliance Control}, and \textit{Storage Access}. Tasks in this tier use fully explicit instructions, such as \textit{``Pick up the apple and place it on the table''}, and test whether the agent can ground named objects, select affordance-compatible actions, and execute short interaction sequences. Strong performance on this tier is necessary but not sufficient: failures here indicate grounding or action-selection problems before higher-level reasoning is required.

    \item \textbf{Tier 2: Reasoning Ability (90 tasks).}
The Reasoning tier contains three scenarios: \textit{Hygiene \& Comfort}, \textit{Food \& Drink}, and \textit{Environmental Regulation}. Rather than explicit commands, agents receive vague human-centric utterances---for example, \textit{``I'm tired''} implying find a bed, or \textit{``It smells in here''} implying locate a trash can. To succeed, the agent must map an implicit human need to a physically grounded action sequence using commonsense knowledge, without the instruction explicitly naming the target object. All Reasoning tasks are single-room by design, so performance gaps on this tier primarily reflect intent resolution rather than cross-room navigation.

    \item \textbf{Tier 3: Long-horizon Ability (90 tasks).}
The Long-horizon tier contains three scenarios: \textit{Study \& Work Setup}, \textit{Entertainment Setup}, and \textit{Home Organization}. Tasks in this tier require coordinating \textbf{three or more objects} over up to \textbf{60 steps}, often involving multiple state changes and object placements. These tasks compound object tracking, spatial memory maintenance, and sequential sub-goal scheduling. Since 45.6\% of Long-horizon tasks require cross-room navigation, this tier is sensitive to both memory retention failures and planning errors; \prism's modular design then makes these factors analyzable through controlled module substitution. Representative task samples from all three tiers are provided in Appendix~\ref{sec:supp_tasks}.
\end{itemize}

\section{Memory types}
\label{app:memory}

\begin{itemize}
    \item \textbf{Long-term Memory} ($\mathcal{M}_{long}$) stores the full apartment room-object inventory---room names and their contained object types---as a persistent oracle. Providing this as ground truth is a deliberate design choice: it removes perceptual reconstruction errors from the analysis, ensuring that failures in cross-room navigation reflect memory \textit{utilization} rather than memory \textit{construction}.
    \item \textbf{Short-term Memory} ($\mathcal{M}_{short}^{(t)}$) tracks the dynamic execution state: the agent's currently held and highlighted object IDs, a \textit{fixed-length} sliding action history buffer, and the state delta $\Delta s_{t-1}$ from the preceding step. The fixed-length buffer is an important design decision---it keeps per-step context overhead constant regardless of episode length, preventing the context saturation that would otherwise emerge in the later stages of Long-horizon tasks.
\end{itemize}

\section{Action Space and State Transition Mechanics}
\label{sec:supp_action}

Table~\ref{tab:action_spec} provides the complete 21-action specification with preconditions and state deltas.

\begin{table}[h]
\centering
\caption{Complete atomic action specification.}
\label{tab:action_spec}
\resizebox{\linewidth}{!}{%
\begin{tabular}{llp{5.5cm}p{4.5cm}}
\toprule
\textbf{Module} & \textbf{Action} & \textbf{Precondition} & \textbf{State Delta} \\
\midrule
Navigation   & \texttt{move\_forward(dist)}     & Passable path ahead           & Position $+= dist$ \\
Navigation   & \texttt{move\_backward(dist)}    & Passable path behind          & Position $-= dist$ \\
Navigation   & \texttt{move\_left(dist)}        & Passable path left            & Position offset \\
Navigation   & \texttt{move\_right(dist)}       & Passable path right           & Position offset \\
Navigation   & \texttt{turn\_left(deg)}         & ---                           & Yaw $-=$ deg \\
Navigation   & \texttt{turn\_right(deg)}        & ---                           & Yaw $+=$ deg \\
Navigation   & \texttt{look\_up(deg)}           & ---                           & Pitch $-=$ deg \\
Navigation   & \texttt{look\_down(deg)}         & ---                           & Pitch $+=$ deg \\
Navigation   & \texttt{move\_to\_room(room\_id)}  & Room exists in apartment    & Position $\to$ room entry \\
Navigation   & \texttt{move\_forward\_to\_wall} & Passable corridor             & Position $\to$ nearest wall \\
\midrule
Grounding    & \texttt{move\_to(obj\_id)}       & Object visible                & Position $\to$ near obj \\
Grounding    & \texttt{highlight(obj\_id)}      & Object in range               & highlighted\_id $=$ obj \\
\midrule
Interaction  & \texttt{open(obj\_id)}           & \texttt{is\_open=False}, in range  & \texttt{is\_open} $\to$ True \\
Interaction  & \texttt{close(obj\_id)}          & \texttt{is\_open=True}, in range   & \texttt{is\_open} $\to$ False \\
Interaction  & \texttt{turn\_on(obj\_id)}       & \texttt{is\_on=False}, in range    & \texttt{is\_on} $\to$ True \\
Interaction  & \texttt{turn\_off(obj\_id)}      & \texttt{is\_on=True}, in range     & \texttt{is\_on} $\to$ False \\
\midrule
Manipulation & \texttt{pick\_up(obj\_id)}       & holding$=$None, in range      & holding $\to$ obj\_id \\
Manipulation & \texttt{drop\_held}              & holding $\neq$ None           & holding $\to$ None \\
\midrule
Placement    & \texttt{place\_on(surface\_id)}  & holding $\neq$ None, in range & \texttt{on\_top\_of} $\to$ surface\_id \\
\midrule
Liquid       & \texttt{fill(container\_id)}     & Near fluid source             & \texttt{is\_filled} $\to$ True \\
Liquid       & \texttt{pour(target\_id)}        & holding $=$ container, in range & \texttt{is\_filled(target)} $\to$ True \\
\bottomrule
\end{tabular}}
\end{table}

\section{Affordance Ontology}
\label{sec:supp_affordance}

All 79 object classes are assigned exactly one affordance type from a fixed ontology of seven categories (Table~\ref{tab:affordance}).

\begin{table}[h]
\centering
\caption{Affordance ontology with supported actions and representative objects.}
\label{tab:affordance}
\resizebox{\linewidth}{!}{%
\begin{tabular}{llp{4.5cm}p{4cm}}
\toprule
\textbf{Affordance Type} & \textbf{Supported Actions} & \textbf{Representative Objects} & \textbf{Key Constraint} \\
\midrule
\texttt{door/container}  & \texttt{open}, \texttt{close}                & Doors, drawers, refrigerators, ovens, toilet covers & Within 1.5\,m \\
\texttt{switch}          & \texttt{turn\_on}, \texttt{turn\_off}        & Lamps, stoves, computers, TVs, speakers, faucets & Within 1.5\,m \\
\texttt{pickup}          & \texttt{pick\_up}, \texttt{drop\_held}, \texttt{place\_on} & Bottles, pillows, chess pieces, keyboards, mice & Holding slot must be empty \\
\texttt{surface}         & \texttt{place\_on} (as target)               & Tables, dressers, desks, counters, shelves & Must hold a \texttt{pickup} object \\
\texttt{fluid\_source}   & triggers \texttt{fill}                       & Faucets, pitchers & Nearby container required \\
\texttt{container\_fluid}& \texttt{fill}, \texttt{pour}                 & Cups, bowls, jugs & Within range of source \\
\texttt{anchor}          & \texttt{move\_to}, \texttt{highlight} only   & Built-in furniture, walls, fixed fixtures & Not directly manipulable \\
\bottomrule
\end{tabular}}
\end{table}

\section{Extended Ablation Analysis}
\label{sec:supp_ablation}

\subsection{Perception Noise Robustness}

To evaluate robustness to imperfect perception, we test two complementary conditions on GPT-5.2 on the 45-task ablation subset: (1) \textbf{controlled random noise}, where a fraction of visible objects is dropped from $\mathcal{P}_t$ before passing to the Memory module; and (2) \textbf{VLM-based perception}, where a frontier VLM identifies visible object \textit{types} from the egocentric image, which are then mapped to oracle object IDs---all simulator-visible instances of the predicted type are treated as detected. This mapping is intentionally partial-oracle: it isolates the VLM's type-level recognition errors from object-ID disambiguation, providing a conservative lower bound on full VLM perception degradation. Table~\ref{tab:perception_noise} reports results across all conditions.

\begin{table}[h]
\centering
\caption{Impact of perception and memory noise on task performance. Baseline uses oracle perception.}
\label{tab:perception_noise}
\resizebox{0.7\linewidth}{!}{%
\begin{tabular}{llccc}
\toprule
\textbf{Noise Type} & \textbf{Drop Rate}
& \textbf{SR (\%)\,$\uparrow$}
& \textbf{Steps\,$\downarrow$}
& \textbf{TC ($\times10^3$)\,$\downarrow$} \\
\midrule
Oracle (Ours) & 0\%  & 74.0 & 26.6 & 113.1 \\
VLM-based  & 0\%  & 62.0 & 34.0 & 211.8 \\
\hline
Perception noise & 20\% & 72.0 & 26.6 & 108.2 \\
Perception noise & 40\% & 64.0 & 31.0 & 126.6 \\
\hline
Memory noise  & 20\% & 72.0 & 28.3 & 117.9 \\
Memory noise  & 40\% & 68.0 & 29.5 & 117.5 \\
\bottomrule
\end{tabular}}
\end{table}

The VLM-based condition produces the largest overall degradation: SR drops by 12\% (74.0\% $\to$ 62.0\%), Steps increase by 7.4, and TC nearly doubles to 211.8K. This outcome is notably worse than even the 40\% random drop condition ($-$10\% SR), despite the type-to-ID mapping providing partial oracle support. The gap indicates that VLM perception errors are not random but \textit{systematic}---type misidentification and missed detections tend to co-occur on task-critical objects (e.g., failing to detect a \texttt{refrigerator} when the task requires filling a container), causing the Memory module to generate semantically mismatched Target Room Predictions that compound across sub-goals in a way that uniform random drop does not replicate. The TC increase further reflects the downstream effect: deprived of correct object anchors, the planner generates longer exploratory reasoning traces in an attempt to recover, consuming nearly twice the tokens without proportional SR recovery.

Under controlled random noise, both types show initial resilience at 20\% drop ($-$2\% SR for both): under perception noise, the \texttt{only\_show\_when\_near} mechanism allows recovery by navigating closer to dropped objects, while under memory noise the Historical Summary partially compensates by retaining prior object sightings. At 40\%, perception noise causes a sharper $-$10\% SR degradation versus $-$6\% for memory noise, concentrated on Long-horizon tasks where missed objects early in the trajectory compound across sub-goals---confirming that accurate initial object detection is the tighter reliability constraint. Once an object is correctly perceived and logged, the memory architecture tolerates partial information loss more gracefully than it tolerates perceptual gaps at the source.

Taken together, the three conditions reveal a degradation hierarchy: oracle $>$ perception noise $\geq$ memory noise $>$ VLM-based, with the VLM gap driven by systematic rather than stochastic errors. These results validate the modular pipeline design: by controlling perception quality independently of reasoning and memory, researchers can attribute SR changes to specific pipeline stages---which is the core diagnostic value of \prism's architecture and the motivation for reporting oracle-perception results as the primary baseline throughout the main paper.

\subsection{Cross-Module Substitution Analysis}

Table~\ref{tab:modular_evaluation} reports performance when Memory and Planning modules are independently instantiated with different LLM backbones, demonstrating that the modular interface works in practice.

\begin{table}[h]
\centering
\caption{Cross-module substitution results. Each row independently varies the Memory and Planning module backbone. Human baseline included for reference.}
\label{tab:modular_evaluation}
\resizebox{0.75\linewidth}{!}{%
\begin{tabular}{llccc}
\toprule
\textbf{Memory} & \textbf{Planning}
& \textbf{SR (\%)\,$\uparrow$}
& \textbf{Steps\,$\downarrow$}
& \textbf{TC ($\times10^3$)\,$\downarrow$} \\
\midrule
Human        & Human        & 92.0 & 15.3 & ---  \\
\hline
GPT-5.2      & GPT-5.2      & 74.0 & 26.6 & 113.1 \\
GPT-5.2      & Claude Sonnet~4.6 & 72.0 & 24.2 & 100.0 \\
Claude Sonnet~4.6 & GPT-5.2  & 70.0 & 23.0 & 99.7  \\
\bottomrule
\end{tabular}}
\end{table}

Fixing Memory at GPT-5.2 and replacing Planning with Claude Sonnet~4.6 yields a $-$2\% SR drop but $-$13.1K TC, indicating that Claude's more concise planning traces reduce cost without significant accuracy loss. Fixing Planning at GPT-5.2 and replacing Memory with Claude Sonnet~4.6 causes a similar $-$4\% SR drop with comparable TC reduction. The 18-point human--machine gap (92.0\% vs.\ 74.0\%) is concentrated on Reasoning and Long-horizon tasks, with human failures predominantly caused by interaction proximity errors---a qualitatively distinct failure profile from LLM failures (sub-goal ordering errors and destination hallucination).

\section{Implementation Details}
\label{sec:supp_impl}

\begin{table}[h]
\centering
\caption{Implementation hyperparameters.}
\label{tab:hyperparams}
\resizebox{0.8\linewidth}{!}{%
\begin{tabular}{lll}
\toprule
\textbf{Parameter} & \textbf{Value} & \textbf{Notes} \\
\midrule
Short-term memory window $N$ & 10 steps & Fixed across all experiments \\
Max episode steps & 60 & Applied uniformly across all tiers \\
Interaction range & 1.5\,m & Simulator-enforced \\
LLM temperature & 0.0 & Deterministic decoding \\
Parallel simulation instances & 8 & Via Unity ML-Agents \\
API endpoint & OpenRouter & Unified interface for all models \\
\bottomrule
\end{tabular}}
\end{table}

\section{Failure Mode Taxonomy}
\label{sec:supp_failures}

Based on manual inspection of failure cases, we identify three primary failure modes:

\begin{table}[h]
\centering
\caption{Failure mode taxonomy across capability tiers.}
\label{tab:failure_modes}
\resizebox{\linewidth}{!}{%
\begin{tabular}{lp{8cm}l}
\toprule
\textbf{Failure Mode} & \textbf{Description} & \textbf{Dominant Tier} \\
\midrule
\textbf{Semantic Hallucination}
& Agent navigates to a plausible but contextually incorrect location (e.g., sofa instead of bed for \textit{``I'm tired''})
& Reasoning \\[4pt]
\textbf{Exploratory Deadlock}
& Agent enters a repetitive navigation loop across 2--3 rooms, exhausting the step budget without locating the target
& Reasoning, Long-horizon \\[4pt]
\textbf{Context Saturation Collapse}
& Agent correctly completes early sub-goals but fails on later ones as early-trajectory context is displaced from the active context window
& Long-horizon \\
\bottomrule
\end{tabular}}
\end{table}

Context Saturation Collapse is uniquely diagnostic: it manifests as a steep intra-episode SR drop strongly correlated with TC divergence (40--100\% more tokens than successful episodes), which directly motivated the Historical Summary design.

\section{Limitations and Responsible Use}
\label{sec:supp_limitations}

\subsection{Benchmark Limitations}

\textbf{Oracle perception.} All experiments in the main paper adopt ground-truth object lists as the perception input, deliberately isolating reasoning and memory failures from perceptual noise. This is a controlled experimental choice, not an architectural limitation---the standardized perception interface supports any perception model. Perception noise experiments and the VLM-based perception stress test in Appendix~\ref{sec:supp_ablation} show how performance degrades under imperfect observations: SR drops by 10\% under 40\% random object drop and by 12\% under VLM-based perception with oracle type-to-ID mapping. A fully end-to-end perception setting without oracle ID mapping remains future work.

\textbf{Scene scope.} \prism comprises five apartment scenes covering all major household functional zones (bedroom, kitchen, bathroom, living room, study, dining area). Results are consistent across all five layouts (per-apartment SR std = 2.5\%), but generalization to non-household environments (offices, outdoor spaces, industrial settings) is not evaluated.

\textbf{Task generation bias.} Task instructions are generated by a frontier LLM and human-verified for feasibility. This pipeline is scalable but may introduce subtle linguistic biases toward instruction phrasings that are natural for the generator model. All tasks underwent human feasibility verification to mitigate this, but the generator's vocabulary and phrasing preferences are not fully characterized.

\textbf{Simulator asset bias.} The 3D assets are sourced from a fixed commercial library and skew toward contemporary Western household aesthetics. Cultural diversity in furniture styles, room layouts, and household objects is limited.

\textbf{Household scope.} \prism evaluates embodied planning within household environments. Claims about agent performance do not generalize to outdoor navigation, industrial manipulation, or multi-agent coordination without further evaluation.

\subsection{Responsible Release}

\prism is an evaluation benchmark for LLM-based embodied planning research. The release comprises simulator binaries, task JSON files, evaluation scripts, and scene configuration files. The benchmark does not train or fine-tune any model, does not contain personal or sensitive data, and does not introduce misuse risks beyond those inherent to general embodied AI research. All 3D assets are licensed for academic research use; terms are documented in the repository. The LLM APIs used for evaluation (GPT, Claude, Gemini, GLM) are accessed via standard commercial endpoints under their respective terms of service.

\section{Artifact Documentation}
\label{sec:supp_artifact}

\subsection{Release Contents}

The \prism benchmark will be released in a public repository, containing:

\begin{itemize}
    \item \textbf{Simulator binary}: Pre-built Unity~6/HDRP executable for Linux/Windows, supporting headless parallel evaluation via Unity ML-Agents.
    \item \textbf{Task dataset}: 300 verified tasks in JSON format, each specifying tier, scenario, apartment ID, task instruction, required actions, initial object states, and cross-room label.
    \item \textbf{Evaluation scripts}: Python API for connecting LLM backends (OpenRouter-compatible) to the simulator, running full evaluation loops, and computing SR/Steps/TC metrics.
    \item \textbf{Scene configurations}: Per-apartment scene graph files encoding room topology, object placement, and affordance annotations.
    \item \textbf{Prompt templates}: Memory module and Planning module prompt templates used in all reported experiments.
\end{itemize}

\subsection{Reproducibility}

All reported results are fully reproducible given the released simulator binary and task dataset. The simulator uses a deterministic physics engine: identical action sequences produce identical outcomes across runs and machines. LLM temperature is set to 0.0 for all experiments (Table~\ref{tab:hyperparams}), eliminating sampling variance. The only non-deterministic component is the LLM API itself; results may vary slightly if model weights are updated by the provider.

\subsection{License}

Benchmark code and evaluation scripts: MIT License. Task dataset (JSON): CC BY 4.0. Simulator binary and 3D scene assets: available for academic research use; commercial use requires separate licensing of the underlying 3D assets. Full license documentation is included in the repository.

\subsection{This Is an Executable Environment, Not a Static Dataset}

\prism is an \textit{executable simulation environment} rather than a static dataset, and therefore does not conform to the Croissant metadata schema, which is designed for static data collections. The benchmark's primary artifact is an interactive Unity simulator that accepts agent actions and returns observations, states, and success signals in real time. Reproducibility is guaranteed through deterministic physics and fixed task specifications rather than through data archival.

\section{Prompt Design}
\label{sec:supp_prompt}

Full prompts for Memory and Planning modules are provided below, along with example inputs and outputs.

\subsection{Memory Module Prompt Structure}

\textbf{Inputs:} Long-term Memory ($\mathcal{M}_{long}$): complete room-object inventory; Short-term Memory ($\mathcal{M}_{short}$): recent $N$-step action history with state deltas and current held/highlighted object IDs; Perception Output ($\mathcal{P}_t$): current visible objects and states.

\textbf{Outputs:} Historical Summary (free-form consolidation of key observations and progress) and Target Room Prediction (inferred next destination).

The Memory module prompt instructs the LLM to output JSON only with keys \texttt{"memory"} and \texttt{"target\_room"}, with explicit rules to avoid action planning (pure memory/navigation summarization only). An example exchange is shown below where the agent has just picked up a toothbrush and must identify the next step for the task \textit{``I just woke up and my mouth feels gross.''}

\subsection{Planning Module Prompt Structure}

\textbf{Inputs:} Task instruction, Memory module outputs (Historical Summary + Target Room), current Room Context (room name + object-type list + adjacent rooms), Visible Objects (unique IDs + states + proximity flags), Room Options (valid room labels for \texttt{move\_to\_room}).

\textbf{Output:} Exactly one atomic action per step in JSON format: \texttt{\{"action": "<name>", "target": "<id\_or\_room>"\}}.

Critical grounding rules enforced at prompt level: (1) object-targeted actions must use exact object IDs from the Visible Objects list---no invented IDs; (2) \texttt{move\_to\_room} target must be an exact room label from Room Options; (3) proximity-gated actions (\texttt{open}, \texttt{pick\_up}, \texttt{place\_on}, etc.) require \texttt{near=true} for the target.

\section{Task Sample Tables}
\label{sec:supp_tasks}

Tables~\ref{tab:basic_tasks}--\ref{tab:lh_tasks} show representative samples from the 45-task ablation subset.

\begin{table}[h]
\centering
\caption{Representative sample of Basic Ability tasks.}
\label{tab:basic_tasks}
\small
\resizebox{\linewidth}{!}{%
\begin{tabular}{@{}cllp{8.5cm}c@{}}
\toprule
\textbf{Task ID} & \textbf{Tier} & \textbf{Scenario} & \textbf{Instruction} & \textbf{Cross-Room} \\
\midrule
B-1-10 & Basic & Navigation \& Inspection & Navigate to the \texttt{white\_queen} first, then go inspect the \texttt{bar\_cabinet}. & Yes \\
B-1-19 & Basic & Navigation \& Inspection & Walk to the \texttt{range\_hood} to observe it, then head to the \texttt{couch} and examine it. & Yes \\
B-2-06 & Basic & Simple Manipulation & Collect the \texttt{white\_king} and arrange it on the \texttt{dining\_table\_top}. & Yes \\
B-2-19 & Basic & Simple Manipulation & Pick up the \texttt{mouse} and relocate it to the \texttt{dresser}. & Yes \\
B-3-10 & Basic & Appliance Control & Switch on the \texttt{table\_lamp} and then switch off the \texttt{floor\_lamp}. & Yes \\
B-3-29 & Basic & Appliance Control & First turn on the \texttt{faucet}, then turn off the \texttt{coffee\_maker}. & No \\
B-4-09 & Basic & Storage Access & Open the \texttt{toilet\_cover} to check inside, then close the \texttt{apartment\_door}. & Yes \\
B-4-30 & Basic & Storage Access & First open the \texttt{apartment\_door}, then go close the \texttt{oven\_door}. & Yes \\
\bottomrule
\end{tabular}}
\end{table}

\begin{table}[h]
\centering
\caption{Representative sample of Reasoning Ability tasks.}
\label{tab:reasoning_tasks}
\small
\resizebox{\linewidth}{!}{%
\begin{tabular}{@{}cllp{8.5cm}c@{}}
\toprule
\textbf{Task ID} & \textbf{Tier} & \textbf{Scenario} & \textbf{Instruction} & \textbf{Cross-Room} \\
\midrule
R-5-04 & Reasoning & Hygiene \& Comfort & I need to use the restroom urgently. & No \\
R-5-14 & Reasoning & Hygiene \& Comfort & I just woke up and my mouth feels gross. I need to freshen up. & No \\
R-5-40 & Reasoning & Hygiene \& Comfort & I feel sweaty and grimy. I need to clean myself up. & No \\
R-6-10 & Reasoning & Food \& Drink & I'm parched and need a glass of water. & No \\
R-6-19 & Reasoning & Food \& Drink & I'd like a warm cup of coffee to start the day. & No \\
R-6-35 & Reasoning & Food \& Drink & I should make a smoothie for a quick energy boost. & No \\
R-7-12 & Reasoning & Environmental Regulation & The room is really dim. I need more light. & No \\
R-7-39 & Reasoning & Environmental Regulation & Someone left the \texttt{oven} door open. I need to close it for safety. & No \\
\bottomrule
\end{tabular}}
\end{table}

\begin{table}[h]
\centering
\caption{Representative sample of Long-Horizon Ability tasks.}
\label{tab:lh_tasks}
\small
\resizebox{\linewidth}{!}{%
\begin{tabular}{@{}cllp{8.5cm}c@{}}
\toprule
\textbf{Task ID} & \textbf{Tier} & \textbf{Scenario} & \textbf{Instruction} & \textbf{Cross-Room} \\
\midrule
L-8-06 & LH & Study \& Work Setup & Get ready for a late-night study session: turn on the \texttt{ceiling\_lamp}, power on the \texttt{computer}, and bring a \texttt{pillow} to the desk chair. & Yes \\
L-8-27 & LH & Study \& Work Setup & Get the workspace ready: turn on the \texttt{ceiling\_lamp} for ambient light, start the \texttt{computer}, and place the \texttt{keyboard} on the \texttt{desk}. & No \\
L-9-05 & LH & Entertainment Setup & Get the living room ready for a party: turn on the \texttt{tv}, turn on the \texttt{speaker}, and place a \texttt{wine\_bottle} on the \texttt{coffee\_table}. & No \\
L-9-25 & LH & Entertainment Setup & Prepare to watch TV: turn on the \texttt{tv}, switch on the \texttt{speaker}, and place a \texttt{candle} on the \texttt{end\_table} for ambiance. & Yes \\
L-10-08 & LH & Home Organization & Do a thorough tidy-up: place the \texttt{pillow} on the \texttt{end\_table}, the \texttt{mouse} on the \texttt{desk}, and close the \texttt{toilet\_cover}. & Yes \\
L-10-26 & LH & Home Organization & Do a thorough tidy-up: place the \texttt{pillow} on the \texttt{dresser}, the \texttt{white\_bishop} on the \texttt{counter}, and close the \texttt{oven\_door}. & Yes \\
\bottomrule
\end{tabular}}
\end{table}


\end{document}